\pdfoutput=1

\documentclass[11pt]{article}

\usepackage[]{EMNLP2023}

\usepackage{times}
\usepackage{latexsym}
\usepackage{graphicx}
\usepackage{array, makecell}
\usepackage{algorithm}
\usepackage{algpseudocode}
\usepackage{comment}

\usepackage[T1]{fontenc}

\usepackage[utf8]{inputenc}

\usepackage{microtype}

\usepackage{inconsolata}

\title{Enhancing Interpretability using Human Similarity Judgements to Prune Word Embeddings}

\author{Natalia Flechas Manrique$^1$, Wanqian Bao$^1$, Aurelie Herbelot$^2$, Uri Hasson$^{1*}$  \\ 
         $^1$University of Trento \enspace
         $^2$Denotation UG 
         \\  \texttt{$^{*}$uri.hasson@unitn.it}\\ }

\begin{document}
\maketitle
\begin{abstract}

Interpretability methods in NLP aim to provide insights into the semantics underlying specific system architectures.  Focusing on word embeddings, we present a supervised-learning method that, for a given domain (e.g., sports, professions), identifies a subset of model features that strongly improve prediction of human similarity judgments. We show this method keeps only 20-40\% of the original  embeddings, for 8 independent semantic domains, and that it retains different feature sets across domains.  We then present two approaches for interpreting the semantics of the retained features.  The first obtains the scores of the domain words (co-hyponyms) on the first principal component of the retained embeddings, and extracts terms whose co-occurrence with the co-hyponyms tracks these scores’ profile. This analysis reveals that humans differentiate e.g. sports based on how gender-inclusive and international they are.  The second approach uses the retained sets as variables in a probing task that predicts values along 65 semantically annotated dimensions for a dataset of 535 words. The features retained for professions are best at predicting cognitive, emotional and social dimensions, whereas features retained for fruits or vegetables best predict the gustation (taste) dimension.  We discuss implications for alignment between AI systems and human knowledge.
\end{abstract}

\section{Introduction}

The popularity of Large Language Models (LLMs) such as ChatGPT\footnote{\url{https://openai.com/chatgpt}} or BLOOM \cite{scao2022bloom} has recently prompted an active area of research around the notion of `alignment', i.e. the ability of NLP models to meet human expectations (see \citealp{wang2023aligning} for a survey). While techniques such as Supervised Fine-Tuning (SFT) and Reinforcement Learning with Human Feedback (RLHF) have become de facto standards to steer models towards human behaviour, the structural differences that make models in need of alignment are not fully elucidated. Why is it that NLP systems organise their knowledge the way they do? And which operations might increase their similarity to human cognition? These questions remain unsolved, not only for LLMs but also for simple word embedding models such as GloVe \cite{Pennington:2014} or Word2Vec \cite{mikolov2013efficient}.

Alongside the question of alignment, a range of model compression techniques have recently been proposed, including pruning, distillation and quantization \cite{xu2023survey}, to increase system efficiency at runtime. Many distilled models perform on a par with their larger counterparts \cite{sanh2019distilbert, jiao2020tinybert}, prompting questions about the nature of semantic encoding in both the original and the compressed architecture. Some investigations have focused on the increased (or decreased) fairness of distilled models, in particular their ability to faithfully reproduce reality, with inconclusive results so far \citep{ramesh2023comparative}. Others have concentrated instead on the correlation between compression and a model's ability to reproduce human behaviour itself \citep{Tarigopula2021.07.08.451521}. Most importantly, there is evidence that pruned networks develop different internal representations \cite{Ansuini_Medvet_Pellegrino_Zullich_2020}. 

In this paper, we bring together the question of alignment and the methodological toolbox given by pruning techniques. The general aim of the following experiments is to understand the semantics of non-contextual word embeddings (GloVe) by evaluating how those embeddings can be fine-tuned in a way that supports explainability and best predicts Human Similarity Judgments (HSJs) for words in specific categories (i.e., co-hyponyms). 

The guiding assumption is that for co-hyponyms that belong to a basic-level category \cite{rosch1976basic}, learning a small subset of relevant features can markedly improve the prediction of HSJs as compared to the use of all features. For example, considering a set of co-hyponyms that all belong to the \textsc{sports} category, it is expected that improved prediction of HSJs between those words can be achieved by identifying a humanly-salient low-dimensional subspace that encodes domain-specific discriminatory properties such as \textit{ball game} or \textit{played in a team}.

We have three aims in this study. Aim 1 is to determine whether pruning improves prediction of HSJs for word-pairs in a study consisting of 8 internal replications (independent datasets). This preliminary step is a necessary prerequisite to support our two subsequent goals, which are focused solely on explainability. Aim 2 focuses on quantifying the position of each co-hyponym in the pruned latent space, and then querying the entire vocabulary to identify words whose co-occurrence with co-hyponyms tracks those positions. This provides a data-driven description of the latent dimensions underlying the pruned feature-set. Aim 3 is to identify the semantics of the pruned sets via a probing task. Specifically, we evaluate how well these (sub)sets of features predict a set of human annotations for a set of 535 pre-defined words for which annotations on interpretable features have been collected \cite{binder2016toward}. 

Our main contribution is in showing that pruning supervised by HSJ is a transparent and effective method to study which human-relevant semantics are contained in word embeddings.


\section{Related Work}

Our work builds upon existing research that establishes a connection between human comparison processes and image representations created by deep neural networks (DNNs). Several prior studies in the area of computer vision have used feature-reweighting \cite{Kaniuth2021FeaturereweightedRS, Peterson2017EvaluatingI} or feature-pruning \cite{Tarigopula2021.07.08.451521}  to improve alignment between human similarity judgments (HSJ) and DNN-generated image representations. Extending the approach to language, \citet{RandB:2020} applied a feature reweighting to optimize the prediction of HSJ from  word-pair embeddings. Following \citet{Peterson2017EvaluatingI} they modeled word-pair similarity as a weighted dot product, using regression, and solving for as many weights as features. While interesting, this procedure has various issues. First, the weights are proportional to feature-products rather than to feature value. This makes the method less interpretable. Second, the method operationalizes the assumption that the DNN has learned a meaningful basis set of features that is applicable to all domains, and that the features just need saliency adjustment per domain. Third, the data do not lend themselves to downstream analyses as they do not select a subset of features or directly reweight them.  
In contrast, the view we present for adapting the features to the domain is that there exists a subspace of meaning/features in the DNN whose saliency is already properly calibrated, and what needs to be done is just to identify the relevant/irrelevant features for the domain. Pruning and reweighting are therefore two different approaches to understanding latent content. 

Prior work has also used pruning to predict HSJ. \citet{Tarigopula2021.07.08.451521} showed that when applied to image embeddings extracted from the penultimate layer of VGG-19, pruning markedly improves prediction of out of sample similarity judgments while maintaining around 20\% or fewer of the layer's features.  However, image and word embeddings are derived in different ways and and it is unclear whether the findings generalize to the text domain. 

The other work relevant to the current effort is on interpretation of word embeddings. \citet{Chersoni2021DecodingWE} and \citet{utsumi2020exploring} present a probing method for studying the semantic dimensions latent in word embeddings by constructing a mapping between embedding-vectors and human-rated semantic features.  We do the same, but instead of using the entire feature set, use the feature-subsets produced by a supervised-pruning procedure. This a fundamental departure from prior work as it ultimately probes for semantic features in pruned embedding subsets and in this way highlights the latent dimensions that are important to humans. 

Finally, our effort links up to recent work on alignment, which focuses on elucidating differences between computational systems and human behaviour: recent examples are \citet{hu-etal-2023-fine}, who compare pragmatic phenomena in humans and Large Language Models, or \citet{bao-etal-2023-human}, who attempt to reproduce human word acquisition by implementing conceptual attribute comparison in their model. Of relevance to our approach, \citet{park2023training} remove spurious correlations between network features via pruning, with the aim to reduce machine-specific biases in the learned model. Their work is however focused on images rather than text.

\section{Datasets}
\label{subsection:description_data}

\subsection{Human similarity judgements}
\label{subsubsection:hsj}

Our HSJ dataset was made available by \citet{RandB:2020} via OSF\footnote{\url{https://osf.io/d7fm2/?view_only=c5ba5d34a5e34ff3970a652c07aadc5c}}. The data covers words in eight categories: furniture, clothing, birds, vegetables, sports, vehicles, fruits and professions. Human similarity ratings were obtained for word pairs within but not across categories, with each category containing around 20-30 words. Participants ($N=365$) were recruited from the US population using an online recruitment and data collection system (mean age = 33 years, 55\% female). Each participant was randomly assigned to one of the eight categories. For most categories, participants only completed some of all possible pairwise similarity judgements, which only made group-level analyses  possible.  Consequently, judgements were averaged across participants and organized in similarity matrices, which we will refer to as representation-similarity matrices (RSMs).

\subsection{Word embeddings}
\label{susubsection:word_embeddings}

For all words in the eight categories, we collected 300-dimensional GloVe embeddings \citep[Global Vectors for Word Representation,][]{Pennington:2014}\footnote{\url{https://nlp.stanford.edu/projects/GloVe/}}. These embeddings are referred to as \textit{GloVe 6b Giga + Wiki} in \citet{RandB:2020}, because they were obtained by training on the GigaWord Corpus and Wikipedia, which have a combined size of 6B tokens. For each of the eight categories, we arranged embeddings into matrices, with words as rows and features as columns. To operationalize word-pair similarity, we computed Pearson's correlations across all embeddings within each category and organized them into RSMs.

\section{Algorithms}

\subsection{Pruning Algorithm and Cross Validation}
\label{subsection:pruning_RB}

Algorithm \ref{alg:cap_1} completely describes our pruning algorithm, which is a sequential feature selection procedure. We briefly summarize its main elements. The objective of the algorithm is to identify a reduced subset of features, so that when that subset is used to produce the $Object \times Object$ Similarity matrix, $SM_{DNNRED}$, the resulting matrix produces a maximal fit to the human similarity judgments. The fit between the two similarity matrices matrices is computed using the Spearman's rank correlation coefficient ($\rho$).  The effectiveness of the pruning solution is evaluated by applying the set of features found for our training data to an unseen test data sample, as explained below.

For both word embeddings and human judgment similarity matrices we create the test and train partitions on a given fold so that the test partition consists of all pairwise similarity ratings associated with a target word $i$. This means that if for $N$ words the number of unique pairwise judgments is $(N^2 - N)/2$, we construct the test partition to consist of the $N$ pairwise similarity judgments associated with the left-out $i$th word. The test partition's size is therefore $(1/N) \times (N^2 - N)/2$, and the train partition consists of all other pairwise judgments. 

As a baseline value, we use the average second order isomorphism measure (2OI) for the test partition for each fold before pruning, as reported in Table \ref{Table:t-test-results}. This was defined as the Spearman's $\rho$ between two sets of similarity matrices: 1) the $N$ similarity judgments associated with the target word as estimated from the full, non-pruned word embeddings, and 2) the ground-truth judgments as provided by humans. The 2OI measure was chosen to make the results comparable with the work of \citet{RandB:2020}.

\begin{algorithm*}[tb] 
 \caption{Pruning: Main algorithm. $\rho$ refers to Spearman's correlation}\label{alg:cap_1}
\begin{algorithmic}\\
\textbf{Inputs}: \\
 \begin{itemize}
    \item $\mathit{SM_{HM}}$: similarity Matrix of human similarity judgments
    \item $\mathit{SM_{DNN}}$: similarity Matrix of similarity estimations derived from the GloVe by computing Pearson's R between the embeddings of two words 
\end{itemize}\\
\break
\\
\begin{enumerate}
    \item \textbf{Compute baseline Spearman's Rho } $\mathit{\rho(SM_{HM}, SM_{DNN})}$, using the full set of features.
    \item \textbf{Rank features}
    \begin{itemize}
        \item For each feature:
        \begin{itemize}
            \item Remove the feature from original embeddings, compute reduced similarity matrix $SM_{DNNRED}$.
            \item Calculate difference $D = \rho(SM_{HM}, SM_{DNN}) - \rho(SM_{HM}, SM_{DNNRED})$.
        \end{itemize}
        \item Rank features based on $D$, with higher values indicating greater importance.
    \end{itemize}

     \item \textbf{Construct pruned embeddings} \begin{itemize}
        \item Initialize an empty set of features.
        \item Iterate over ranked features in descending order of importance according to $D$
        \begin{itemize}
            \item Reinsert one feature at a time.
            \item Calculate $\rho$ after each feature reinsertion, store values in array $a$.
        \end{itemize}
        \item Determine the maximum value in array $\mathit{a}$. 
        \item The index of the maximum value  delimits the set of features to be included in the pruned embeddings.
     \end{itemize} 
 \end{enumerate}
 \end{algorithmic}
\end{algorithm*}

\subsection{Feature set interpretation}
\label{sec:interpretation-algorithm}

After applying the pruning algorithm, we perform a Principal Components Analysis (PCA) and identify those vocabulary words whose co-occurrence profile with the category words tracks the first-PC scores for those words. This results in a human-readable representation of the main discriminative features in the pruned space. The process is achieved in two steps, as detailed below. As a control we also applied this PCA analysis to non-pruned embeddings 

\subsubsection{Identifying a word's immediate context}

To compute the PMI of each vocabulary word with each of the category words we used  code provided by \citet{KandH:2021}\footnote{\url{https://github.com/akb89/counterix/blob/master/counterix/core/weigher.py}} who computed the Positive PMI between all word combination in the WIKI4 corpus (4\% of the English Wikipedia sampled across the entire dump). 

We use PMI rather than positive PMI (PPMI) because, for our analysis, the extent to which pairs of words co-occur less frequently than would be expected by independence is also meaningful.

Note that for purposes of the current analyses, we needed to identify those vocabulary words that were part of the contexts for each of the words in the category. This category-related corpus vocabulary is created as follows: For every word in a given category (target word), we select words whose joint probability with the target word is not zero (i.e. words whose PMI value with the target word is not zero), forming a set of context words for a given target word. More formally, we denote the set of all vocabulary words as $V$. We can define an ``immediate context'' subset, denoted by $N(i)$, as the subset of words in $V$ that are adjacent to the target word $i$, within a word window of $\pm2$ words:
$N(i) = \{w \in V | P(w, i) \neq 0\} $
\\
where $w$ is an element (word) in $V$, and $P(w, i)$ denotes the joint probability of word w and target word $i$.

After computing the immediate contexts for each category co-hyponym, we combine these context-sets. We denote the category as $C$ and the vocabulary as $V$. The "Category context" or the vocabulary of the category is the union of the context word sets of all the target words in the category, $C = \bigcup N(i)$. This combines, for a given category, all the immediate-contexts subsets.

\subsubsection{Find the correlations between the PMI vectors and the first PC of each category}
\label{sec:PC1correl}

For each category, the Spearman’s correlation between each PMI vector of each corpus-vocabulary word and the category-words' scores on the first principal component was calculated, alongside its statistical significance. The vocabulary was ranked depending on the correlation results.

Many vocabulary words end up in the Category context-set even though they are not lexical items relevant to our analysis. For this reason we only included words that were within the most frequent 15K dictionary-words in the corpus, and further eliminated proper nouns and numbers. As relevant correlations we considered those whose statistical significance satisfied $p < .05$. Finally, we required that more than $60\%$ of the components of the PMI vectors must be non-zero values so that the correlation was not driven by a few zero vs non-zero entries.

\section{Results}
\subsection{Improved prediction of human judgments}

\begin{table*}[tb]
    \centering
    \begin{tabular}{lllll}
        \textbf{Category} & \textbf{Baseline Mean} & \textbf{Pruned Mean} & \textbf{T value} \textbf{(Pruned-Baseline)} &  \textbf{Features Retained} \\ \hline
        \textbf{Furniture} & 0.46 (0.19) & 0.63 (0.25) & 4.47*** &  121.00 (19.72) \\ 
        \textbf{Clothing} & 0.37 (0.16) & 0.52 (0.21) & 4.74*** &  84.21 (11.99) \\ 
        \textbf{Vegetables} & 0.30 (0.28) & 0.45 (0.30) & 3.59** &  58.05 (26.22) \\ 
        \textbf{Sports} & 0.40 (0.19) & 0.52 (0.20) & 4.13*** &  101.39 (16.84) \\ 
        \textbf{Vehicles} & 0.66 (0.12) & 0.74 (0.15) & 3.78** &  131.05 (23.84) \\ 
        \textbf{Fruit} & 0.38 (0.24) & 0.42 (0.26) & 0.66 &  88.48 (16.16) \\ 
        \textbf{Birds} & 0.20 (0.14) & 0.37 (0.25) & 3.58** &  57.57 (13.23) \\ 
        \textbf{Professions} & 0.45 (0.20) & 0.57 (0.18) & 3.72*** & 102.43 (9.96) \\ 
    \end{tabular}
    \caption{Prediction accuracy (Spearman's Rho) for human similarity judgments from GloVe embeddings. Baseline: prediction for test partition when using all GloVe features. Pruned: predictions based only on the pruned set learned using the training partition. Features Retained: average number of featuers retained from training $\pm SD$. T values are from paired T-tests within category. ** $p < .01$, *** $p < .001$. }
    \label{Table:t-test-results}
\end{table*}

For each fold we compute the baseline 2OI of the test partition, and the 2OI value computed when using the pruned feature-set identified by the algorithm using the train partition. The results are summarized in Table \ref{Table:t-test-results}and show that for all 8 data sets, pruning improved out of sample prediction of human behavior, in some cases by considerable magnitudes. 

Because, within each category, there are as many test-partition folds as words, we could compute a paired T-test between each test partition's baseline Spearman $\rho$ (prior to pruning), and the $\rho$ value obtained after pruning.
As shown in Table \ref{Table:t-test-results}, the difference in 2OI values was statistically significant for 7 of the 8 datasets.

Further, the number of features retained through pruning varied considerably across categories.
Notably, within each category, the standard deviation of this statistic across folds was low, meaning that pruning produced relatively systematic set-sizes for different train folds. 

\subsection{Supervised-pruning selects for different features across domains}
To determine whether a core set of features was maintained across the eight categories, we simply summed, for each GloVe feature, the number of times it was retained for each of the 8 prunings. We found that no feature was retained across all 8 datasets or even 7 of the 8.  The strongest overlap was seen in 6 features that were included in 6 pruned datasets. However, 220 of the features were kept for only 3 datasets or less. Thus, there was no core set of features that remained in all cases. 

\begin{figure}[tb]
\includegraphics[width=0.5\textwidth]{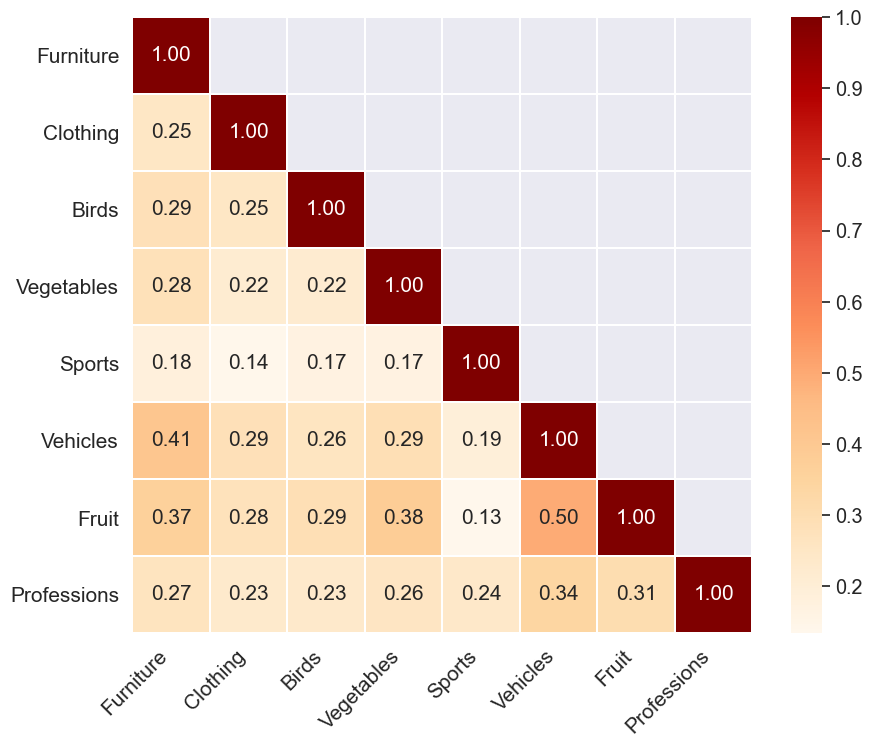}
\centering
\caption{Dice coefficient indicating overlap of features sets pruned by different categories}
\label{fig:DiceRB}
\end{figure}

We also evaluated whether there were category pairs which, when pruned, tended to maintain similar sets of features, which would be an indicator of similar semantics. For each pair of categories we computed the Dice coefficient between the two sets \cite{dice1945measures}.  
As can be seen in Figure \ref{fig:DiceRB}, the value of the coefficient was low across the board, and most so for \textsc{sports}. 

\subsection{Pruned feature sets are interpretable}
\label{section:PMI}

To interpret the semantics of the pruned feature sets we applied PCA to the pruned embeddings and interpreted the results as detailed in \S\ref{sec:interpretation-algorithm}. In this analysis we did not use cross validation, but pruned the complete word-by-features embedding matrix for each of the eight datasets.  For example, if 120 features were retained by pruning the \textsc{furniture} embeddings, we applied PCA to the 20 (furniture words) x 120 (retained features) embedding matrix. 

\begin{table}[]
    \centering
    \begin{tabular}{lllll}
    \hline
        \textbf{Category} & \textbf{Hits} & \textbf{Prn} & \textbf{Full} & \textbf{Cmn} \\ \hline
        \textbf{Furniture} & 15968 & 45 & 23 & 19 \\ 
        \textbf{Clothing} & 8615 & 20 & 11 & 6 \\ 
        \textbf{Birds} & 8850 & 12 & 10 & 6 \\ 
        \textbf{Vegetables} & 2888 & 1 & 2 & 0 \\ 
        \textbf{Sports} & 16075 & 92 & 99 & 58 \\ 
        \textbf{Vehicles} & 18146 & 22 & 44 & 0 \\ 
        \textbf{Fruit} & 8263 & 4 & 6 & 2 \\ 
        \textbf{Professions} & 25125 & 92 & 127 & 88 \\ \hline
    \end{tabular}
    \caption{Hits: Size of Category context-set. Prn/Full: number of words  significantly correlated with the scores of the category’s first PC when computed from Pruned or Full (unpruned) embeddings. Cmn:  number of words in common for pruned and full solutions.}
    \label{tab:PMI_hits}
\end{table}

Table \ref{tab:PMI_hits} presents the sizes of the Category contexts per category (Hits) and of those, the number of words whose PMI correlated significantly with the scores on each category's first PC. We can observe the number of words showing significant correlates was relatively similar when using pruned and non-pruned embeddings. However, the overlap was not necessarily strong in all eight domains.

\begin{figure*}[tb]
\includegraphics[width=1\textwidth]{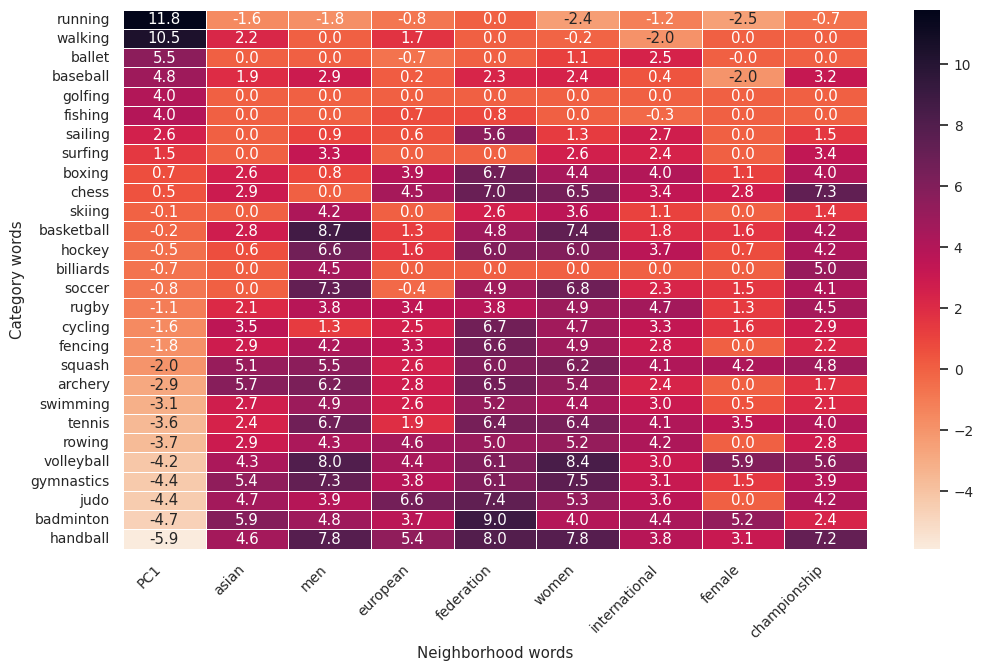}
\centering
\caption{PMI values for words that correlate with co-hyponym scores on first PC computed from pruned embeddings}
\label{fig:PMIvals}
\end{figure*}

To understand how meaning is organized in the pruned embeddings, the first evaluation is based on examining the scores of the category-co-hyponyms on the first Principal Component.  These are shown, for \textsc{sports}, in the first column of Figure \ref{fig:PMIvals}.  One can immediately see a separation between more typical team sports which are here are associated with negative-sign scores, and less typical sports, including running, walking and ballet.  

For \textsc{sports},  the top-20 correlated words include \textit{asian, men's, european, federation, women's, international, female,  championship}, see full results here in attached file. These align with the 1st PC scores in having high PMI with words in the \textsc{sports} category such as basketball, tennis, gymnastics and soccer and low PMI with ballet, golfing, fishing or chess. This list of words returned by the query emphasizes the international and inclusive (gender, country) dimension of some sports vs. others. This of course does not mean that running and walking are less federated or international than the others, it just means that these concepts are less frequently associated when discussing these sports. 

We also find meaningful divergences between the words identified as correlated with the 1st PC of the pruned and non-pruned embeddings as this indicates differences emphasized via pruning. For \textsc{sports}, the noun \textit{player} is identified for the pruned embeddings. In contrast, for the non-pruned embeddings, the verb \textit{play} is more dominant, as well as its morphological variations  \textit{played, playing, plays}.  This appears to emphasize the competitive/non-competitive dimension which was not as salient in the pruned embeddings.  In contrast, the pruned results include \texttt{olympic} and \texttt{medal} whereas the nonpruned do not. 

In \textsc{furniture}, the selected words under the pruning condition appear to highlight spatial and physical dimensions, including prepositions and modifiers such as \textit{out, center, around}. 
For the unpruned condition, on the other hand, words associated with technology are more prominent (\textit{technology, system, powered}. Interesting divergences were found also for several other categories. For example, for  \textsc{clothing}, the pruned embeddings more strongly emphasized the condition of clothes as being new or worn (e.g., \textit{worn, wearing, wore, new}). In contrast, for \textsc{vehicles} the full embeddings emphasized more strongly the verb \textit{drive} and its morphological variants including \textit{drives, driver, driving, drive, driven, drove}). Thus, the dimension of being driven is fleshed out when analyzed against the full embeddings, but not against the pruned ones. Because human comparisons are not strongly based on this dimension, it is effectively partialed out via the supervised pruning. This shows how a dimension may be central within text-meaning (corpus) but not human meaning.

\subsection{Pruned feature sets predict basic semantic features}
\label{section:binder_features}

\begin{figure*}[tb]
\includegraphics[width=0.9\textwidth]{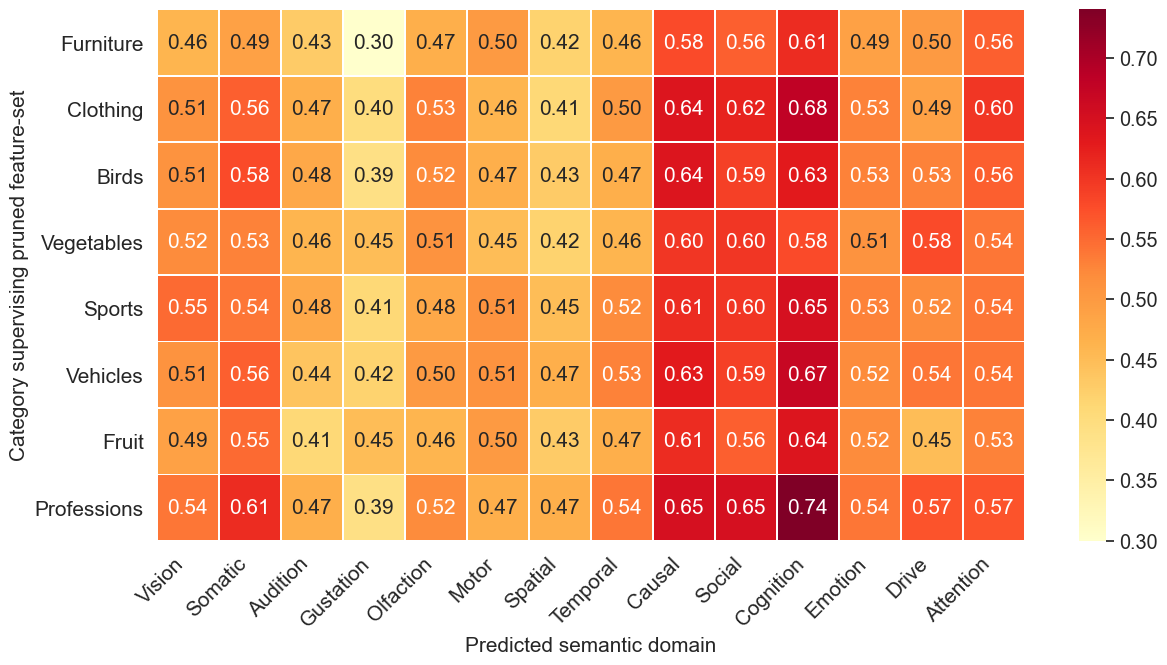}
\centering
\caption{Accuracy in predicting human ratings using pruned features. Cell-values indicate the Spearman's $\rho$ between PLSR predicted results from pruned embeddings ($N=60$ features per category) and the actual values from Binder's dataset. }
\label{fig:Binder60}
\end{figure*}

Having shown that our pruned feature sets can be given a human-readable interpretation, we turn to our final question and seek to explain why they provide better correlation with HSJs, i.e. \textit{why they align}. To do so, we use the curated dataset of \citet{binder2016toward}, which consists of 535 words with human ratings on 65 semantic features belonging to 14 basic semantic areas including Vision, Gustation, Temporal, Causal, and Cognition. Previous work \cite{Chersoni2021DecodingWE, utsumi2020exploring} constructed a regression model (Partial Least Squares Regression; PLSR) that successfully predicted the 65 dimensions from those words' GloVe embeddings. We use the same procedure, but apply it to the feature sets found via supervised pruning to determine if sets pruned by different domains encode different semantics

For each of the 8 categories, we trained a PLSR model on 534 words.\footnote{\textit{Used} appears twice as noun and verb separately, but in GloVe there is only one \textit{used} vector and thus one fewer word than in Binder's actual dataset.} mapping GloVe to Binder features using each category's pruned features. The trained model was applied to the left-out word, predicting its 65 feature-values (leave-one-out cross validation; LOOCV). This resulted in a $533 \times 65$ prediction matrix for all words. For each of the 65 features we could compare the values in the prediction matrix to the true values, using Spearman's $\rho$. Finally, we averaged the correlation values within the 14 larger-scale semantic areas to obtain a single value per domain. 
Note that for this analysis, we only used the top-60 ranked features for each category, thus ensuring that the model got the same amount of data across experiments for a fair comparison between the different pruned feature sets. We chose the value of 60 as it approximated the number of features retained for Clothing, Birds, Vegetables, Professions, and Sports when applying pruning outside a cross-validation framework.

Figure 3 shows the results. Replicating prior work \cite{Chersoni2021DecodingWE, utsumi2020exploring} for all 8 domains, the Cognition area was generally best predicted, while Gustation and Space were generally predicted less well. The features retained for  \textsc{professions} offer the highest prediction scores for 7 of the 14 semantic areas, with a very large relative advantage for prediction of Cognition-related semantic features, and for features in the Social category. \textsc{professions} also predicted Emotion features best, though with a weaker advantage compared to other category's' pruned sets. In contrast, the pruned feature sets from \textsc{fruits} and \textsc{vegetables} predicted Gustation features the best, whereas \textsc{vehicles} and \textsc{sports} predicted Motor features best. 

Interestingly, feature sets pruned for  \textsc{clothing} produced the best prediction of Olfaction, which consisted of one human rated dimension: ``having a characteristic or defining smell or smells''. The different sorts of clothing materials in the set \{mittens/wool, belt/leather, beanie/cloth\} could have been a relevant dimension for comparison. \textsc{clothing} also produced the best prediction of Attention which consisted of two dimensions, `` someone or something that grabs your attention'' and ``someone or something that makes you feel alert, activated, excited, or keyed up in either a positive or negative way''. The fact that these two dimensions were predicted by clothing may be due to the specific items in the set that differed according to evening-wear/non-eveningwear, intimate/non-intimate items, mens/womens wear and sports/non-sports (e.g., suit/jeans, skirt/pants, pajamas/tuxedo). 

These results suggest that the identified dimensions contained in the pruned sets reflect information that is central to the way people compare objects in these categories.

\section{Conclusion}

We have shown in this work that supervised pruning applied to  a word embedding model can improve prediction of human similarity judgements. The method was shown to select different features across domains, demonstrating high levels of conceptual discrimination. Further, the pruned feature sets were interpretable with respect to the first component of a PCA analysis, allowing us to describe how humans discriminate between elements of a category -- and giving the model itself a tool to justify its semantic `beliefs' to a potential user. The probing task provided additional, fine-grained information on the semantics in the different pruned feature-sets.

Given its performance and inherent interpretability, supervised pruning can advance computations related to word-similarity, word-analogy and sentiment analysis, as well as domain adaptation. Most importantly, this method allows an AI system to construct a model of human domain knowledge through pruned embeddings, which diverges from the AI system's own internal organization of the domain (i.e., via full embeddings). This alignment can increase the synergy between the AI system and users, establishing a more robust foundation of shared understanding.

\newpage

\section*{Limitations}

In Aim 1, one important question that remains to be answered is how the results would generalize if the type of embedding and / or the similarity dataset were different. Furthermore, understanding how this methodology could be applied to contextualized embeddings remains to be explored. 

In Aim 2, we based our analysis on the first principal components of the datasets, another interesting venue would be to analyse the second principal component as well.

In Aim 3, We use leave one word out CV. As indicated by \citet{utsumi2020exploring}, this might be problematic for this dataset because it gives a relative advantage to left-out words that have many semantically-similar words in the training set. \citet{utsumi2020exploring} predicts word-cluster features instead, but the limitation of that method is in determining cluster semantics.  \citet{Chersoni2021DecodingWE} predict both single-word and cluster semantics. The decision to use leave-one-word-out CV may contribute to why some domains (e.g., Cognition) are predicted better than others, but does not contribute to differences in prediction for different pruned feature sets, as in all cases the same 535 words are mapped from GloVe to Binder features. 

Finally, we consider that the dimensions highlighted by supervised pruning may be related to the the set of words being compared, to the extent that the similarity ratings are impacted by contrast-relation in the specific category set.  This is to say that if similarity ratings were obtained for a different set of, say, sports co-hyponyms, the dimensions identified may differ.

\bibliography{anthology,custom}
\bibliographystyle{acl_natbib}




\end{document}